# An Automatic Image Content Retrieval Method for better Mobile Device Display User Experiences

Alessandro Bruno
Department of Computing and Informatics at Bournemouth University
Fern Barrow, Poole, Dorset, BH12 5BB, United Kingdom
abruno@bournemouth.ac.uk

**ABSTRACT**

A growing number of commercially available mobile phones come with integrated high-resolution digital cameras. That enables a new class of dedicated applications to image analysis such as mobile visual search, image cropping, object detection, content-based image retrieval, image classification. In this paper, a new mobile application for image content retrieval and classification for mobile device display is proposed to enrich the visual experience of users. The mobile application can extract a certain number of images based on the content of an image with visual saliency methods aiming at detecting the most critical regions in a given image from a perceptual viewpoint. First, the most critical areas from a perceptual perspective are extracted using the local maxima of a 2D saliency function. Next, a salient region is cropped using the bounding box centred on the local maxima of the thresholded Saliency Map of the image. Then, each image crop feds into an Image Classification system based on SVM and SIFT descriptors to detect the class of object present in the image. ImageNet repository was used as the reference for semantic category classification. Android platform was used to implement the mobile application on a client-server architecture. A mobile client sends the photo taken by the camera to the server, which processes the image and returns the results (image contents such as image crops and related target classes) to the mobile client. The application was run on thousands of pictures and showed encouraging results towards a better user visual experience with mobile displays.

## 1 Introduction

Mobile devices, such as smart-phones, today have become powerful image processing instruments and most users spend many times experiencing display. The mobile terminal increased powerful performance and most smart-phone are equipped with high-resolution camera at present. All this enables a new class of applications that use the camera phone: image cropping, seam carving, image classification, mobile visual search, content-based image retrieval. These applications can be used for identifying products, logos recognition and managing personal photo collection. The focus of this paper is on Image Cropping and Image Classification. More particularly, it is implemented a mobile application that automatically crops the photo and detects the class of the object present in the image (flowers, baby, dog, car, airplane, bicycle, bird, bottle, cat, fish…). Image cropping is a technique that is used to select the most relevant areas of an image and discarding the useless parts. Handmade selection, especially in case of large photo collections, is a time-consuming task. Automatic image cropping techniques may suggest to the user which part of the image is the most relevant, according to specific criteria.

Image Classification involves the identification of objects or the labelling of images with keywords. This helps to bridge the semantic gap [1] because this kind of annotations (keywords and object labels) is near to high-level semantic description for good image retrieval.

There are two main possible architectures for our application: 1) a mobile device processes the query image, extracts and transmits feature data; an image classification algorithm runs on a server, then the results of the crop and classification algorithms are sent to mobile client; 2) a mobile client transmits the query image to the server. The image crop and classification algorithms run entirely on a server that sends the results to the mobile client. When the training set consists of thousands of images it has to be placed on a remote server. It is preferred the second architecture in which the mobile only transmits the query image to the server (the training set consists of 1000 images).

The application mainly consists of four parts: 1) Saliency Map Extraction; 2) Image Crop; 3) Features Extraction; 4) Image Classification. In this application we locate object position in a picture by looking at salient function peaks as centroids of salient regions containing the objects to be cropped. The rest of the paper is organized as it follows: sections 2, 3 and 4 provide the reader with an overview of image crop, visual saliency and image classification, section 5 shows the experimental results, section 6 ends the paper with conclusions and future works.

## 2. Content Based Image Retrieval

In this paragraph, the step of information retrieval from the images treated is highlighted. The main objective is to make the visual experience through mobile displays richer of details and semantic description, such as name of the object in the images and good region crops containing details of each object. To do that, the saliency method proposed by Judd et al. in [21] is used. This is an approach based on low, middle and high-level features from images to detect the most important regions from a visual attention perspective. Now, the reader is invited to imagine user experience scenarios where one wants something summarizing the images the user is looking at with details such as number of objects, object labels and image crops. One wants to extract as much information as possible by using visual attention detection as in [33] where the authors performed Content Based Information Retrieval by using a Saliency algorithm to predict Human Fixation for a given image (see figure 1).

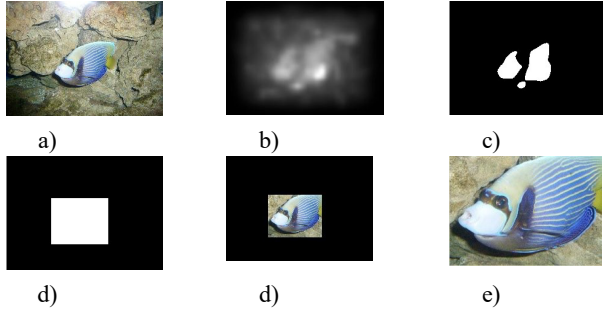

**Figure 1:** The image above is centred with respect to the most important visual object (a) of the scene (a fish). To crop the image, Judd et al. saliency approach (b) is used, then all local maxima of the visual saliency function are extracted (c). Then the bounding box of the region/regions is /are extracted and the crop is given as output with the category label.

## 3. Visual Saliency

Saliency or Visual Saliency deals with identifying the most important regions of an image from a perceptual point of view [11]. The objective of Visual System is to replicate the Human Visual System behaviour. A saliency map is a representation of the salient areas, analysing image properties. Most approaches for visual saliency detection are inspired by HVS (Human Visual System) and tend to replicate the modifications of cortical connectivity for scene perception. Visual Saliency approaches, in scientific literature, can be subdivided into three main groups: Bottom-up, Top-down, Hybrid. Bottom-up approaches (stimulus driven) consider human attention as a cognitive process that selects most unusual aspects of an environment while ignoring more common aspects. In [12] the authors extract various feature maps using the centre-surround differences. In [13] multiscale image features are combined into a single topographical saliency map. A neural network then selects attended locations in order of decreasing saliency. Harel et al. in [14] proposed graph-based activation maps (Graph Based Visual Saliency). Ardizzone et al. [15] proposed SIFT Density Maps (SDM) to study the spatial distribution of SIFT keypoints [16] into the image and its relationship with real human fixation points. In Top-down approaches [17,18] the visual attention process is considered task dependent, and the observer's goal in scene analysis is the reason why a point is fixed rather than others. Object, text and face detection are examples of high-level tasks that guide the human visual system in top-down view. Hybrid approaches [19,20] are combinations of Top-down and Bottom-up methods. In [19,20] the authors use face and text detection as top-down component. A visual saliency approach was proposed by Judd et al. [21] who built a database [22] of eye tracking data from 15 viewers. Low, middle and high-level features of this images have been used to learn a saliency model. Visual saliency approaches are also adopted for IQA (Image Quality Assessment) [34]. In this application, Judd et al. method [21] is adopted because of its accurate performances compared to human fixation points. In greater details, once the saliency map (as in figure 2b) of a given image is extracted (see figure 2a). To get the number N of object blobs contained in the image a threshold to the saliency function (experimentally tuned) is applied, and the number of convex regions is counted (figure 2c). Getting back to the saliency function the local maxima of visual saliency are extracted. Saliency is to be considered as a 2-dimensional function.

After detecting the first local maxima of the saliency function along the image the first crop in the image is extracted.

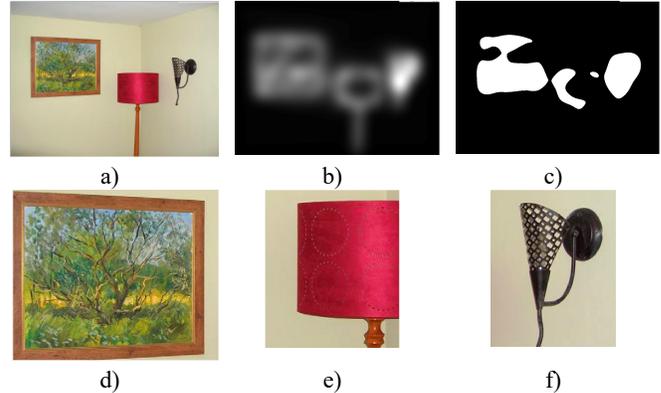

**Figure 2:** The image above is centred on the most important visual object (a) of the scene (a fish). To crop the image, Judd et al. saliency approach is used (b), then all local maxima of the visual saliency function are extracted by looking around the local maxima and the contiguity of the regions treated. Then the bounding box of the region/regions is /are extracted and the crop is given as output with the category label (figures 2d-f).

That is accomplished by first fitting a Gaussian function centred on local maxima in question. The spread of the Gaussian model fitting the saliency blob (around the local maxima) is function of the size of the salient region. Only the most 60% salient pixels along the Gaussian blob were considered for the purpose. The bounding boxes centred on the local maxima of saliency along the image pixels are extracted. After that, all pixels included in the bounding box are not considered anymore as saliency local maxima. Therefore, the process is iterated with the number of N, the number of object blobs in the image.

## 4. Image Classification

The objective of image classification methods, feature based, is to detect the most important and discriminant image features to represent the information content of the image. In [23] the authors, for each image in the database, extract local patches and associated low-level descriptors. In the space of low-level feature, they create a visual vocabulary and then they compute high level image representations for salient and non-salient regions. Each image in the dataset has a signature based on the high-level representation. In the image classification literature, the traditional approach to transform low-level features into high-level representations is the bag-of-visual-words (BOV) [24]. Perronnin and Dance [25] showed that Fisher Kernel outperforms bag-of-words in image categorization scenarios. In [26] the authors performed an image classification approach suitable for training and recognition on a Smartphone. For each of image classes they used 20 training images at moderate resolution, the single class build from different pictures of the same physical object (In the training set of this method the object is always in the foreground and the background shows homogeneous surfaces). The authors extract a features vector composed by colour, edge, interest points, and frequency-based features.

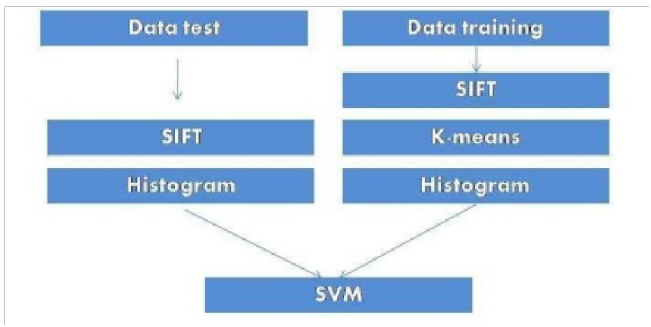

Figure 3: The architecture of the classification system is shown as above. This module allows for extracting each object class from the image the users is looking at.

The authors extract a features vector composed by colour, edge, interest points, and frequency-based features. They used Support Vector Machine for classification. In [31] the authors introduce the notion of discriminative class-specific priors with class specific dictionaries applied in Bayesian sparse regression. More in details, in [31] the proposed framework takes the burden off the demand for abundant training image samples necessary for the success of sparsity-based classification schemes. In [32] the image is classified using multiscale information fusion. In [9] Luo performed image cropping method based on subject detection algorithm, using a belief map that probabilistically indicates the subject content. In [10] Nishiyama et al. build a quality classifier using an image dataset with quality scores. They finally used the classifier to find the cropped region with the highest quality score. In this work the training set is composed by ten image classes organized as in ImageNet repository: flowers, baby, dog, car, painting, bicycle, bird, bottle, lamp, fish, no-class. The training set includes 10 image classes, each of them containing almost 100 sample images (randomly taken from internet) with low, moderate and high spatial resolution. SIFT interest points descriptors (SIFT descriptor is a 128-dimensional vector) are used to create an Image Vocabulary. K-means algorithm clusters the SIFT interest points in k visual words centres. A visual word histogram for each image in the training set is computed, an SVM is trained along the visual word histograms (see figure 3).

## 5. Experimental Results

### 5.1 Training Step
The training set for this system is composed of 1000 images, 100 for each of image classes. The images are taken from [29-30], organized in semantic classes. Every image is cropped with a very simple system: the image crop is the bounding box of the thresholded saliency map of the image. Experimentally, the best threshold value for the application is 0.5. Then, I computed SIFT interest points descriptors for each image (SIFT descriptor is a 128-dimensional vector). SIFT descriptors are used to create an Image Vocabulary. K-means algorithm clusters the SIFT interest points in k visual words centres. The structure of Image Vocabulary has the following two fields:

1. Words (128 x k matrix, where k is the number of interest points clusters);
2. Kdtree (KD-Tree indexing of visual words is used for fast quantization of the image descriptors).

When the image vocabulary is created, visual word histogram for each image in the training set is extracted. SVM [27,28] is adopted for a multiclass problem in which M=10 (the number of classes) and $\omega_{i, i = 1, \ldots, M}$ denote the M class. In a multiclass problem the $i_{th}$ classifier output function is trained by taking the examples from $\omega_i$ as positive and the examples from the other classes as negative. In this case, the $i_{th}$ classifier output function is trained taking the visual word histograms from $\omega_i$ as positive and the visual word histograms from the other classes as negative (one versus all approach). The training phase time is shown in figure 4. It depends on the number of images per class. In table 1 the training accuracy measurements with respect to the number of images per class are shown. The training accuracy percentage grows with the number of class images; it reaches nearly 92% when 100 images are used per each class.

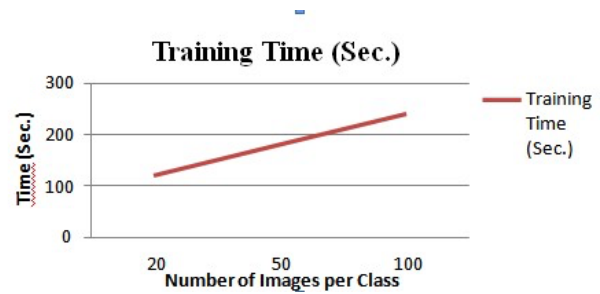

Figure 4: The training time of the classification system is shown above with respect to number of images per class.

Table 1 Test images, number of Images per training class and the training accuracy percentage are shown below.

| Test Images | Number of Images per training class | Accuracy |
|---|---|---|
| 500 | 20 | 72% |
| 500 | 50 | 80% |
| 500 | 100 | 92% |

## 5.2 Testing Step

The Image Classification system includes eleven image categories (flowers, baby, dog, car, painting, bicycle, bird, bottle, lamp, fish, no-class), the category "no-class" involves all the outliers of the SVM predictions. The accuracy is measured as the ratio between the number of correct classifications and the total number of tests. In table 1 the accuracy of the application is shown on a test set of 500 images. The accuracy grows along with the number of Images per Training Set Classes. The mobile application achieves excellent results in terms of accuracy values (92%). The performances are comparable with the method in [26]. Overall, the system reaches an accuracy slightly lower than the method in [26], but it is necessary to mentioned that only SIFT features are used. In [26], the authors use a vector descriptor including: colour information, interest points, frequency descriptors, edge activity. In addition, many test images are taken from the same camera and with well-defined conditions of light (test images are taken from internet with randomly lighting conditions). In tab. 2 Precision and Recall are listed. The results of tests are subdivided in:

1. False Positive (FP) - The image belongs to no-class category but is classified as belonging to a class.
2. False Negative (FN) - The image belongs to a class but is classified as belonging to no-class category.
3. True Positive (TP) - The image is correctly classified.

A result of the method is shown in figure 5; the image is decomposed into 3 image crops, 3 class labels and the counts of the class label instances.

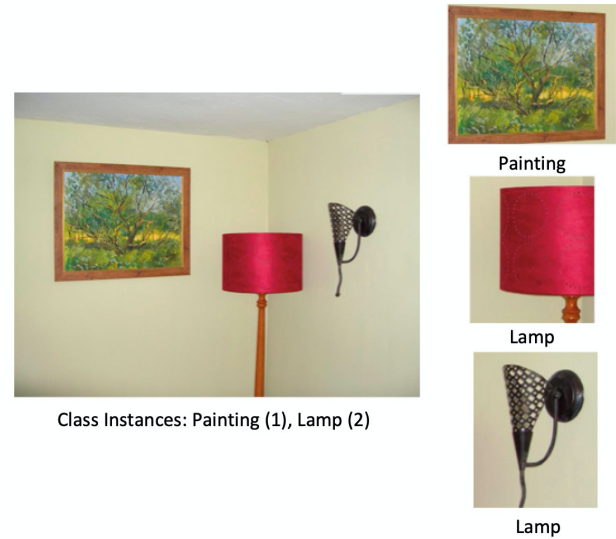

Class Instances: Painting (1), Lamp (2)

**Figure 5:** For a given image the objects are automatically shown with respect to a perceptual viewpoint (they are cropped with a saliency-based method) and a visual category classification is done with SIFT and SVM.

**Table 2 Test images, True Positives (TP), False Positives (FP), False Negatives (FN), Precision and Recall are shown.**

| Test Images | TP | FP | FN | Precision | Recall |
|---|---|---|---|---|---|
| 500 | 460 | 14 (3%) | 26 (5%) | 97% | 95% |

## 6. Conclusions & Future Works

In this paper a new mobile application to enrich the quality of a mobile device display user experience is proposed. In a few words, Saliency is used as a means to crop areas of interest; while an Image Classification module based on SIFT extraction and an SVM retrieve information from the images the user is looking at. The experiments show good performance in terms of precision and recall. The results are comparable with the other techniques of the state of the art. The training set is performed using only visual words of SIFT descriptors.

SIFT descriptors are robust against affine transformations and this gives extra strength to our image classification system. No colour features are used, only SIFT descriptors. Image classification on mobile devices can be used when the user wants to get additional information about the surrounding environment or when the user wants to manage the personal photo album collections. The future work will be focused on extending the training set, in order to recognize a larger number of image categories.